%% file: main.tex
\documentclass[sigconf]{acmart}

\usepackage{enumitem}
\usepackage[mathscr]{eucal}
\usepackage{bm}
\usepackage{algorithm}
\usepackage{subcaption} 
\usepackage{textcomp}
\usepackage{array}         
\NewDocumentCommand{\vect}{ O{} O{} m }{\mathbf{#3}\ifthenelse{\isempty{#1}}{}{^{(#1)}}\ifthenelse{\isempty{#2}}{}{_{#2}}}
\NewDocumentCommand{\mat}{ O{} O{} m }{\mathbf{#3}\ifthenelse{\isempty{#1}}{}{^{(#1)}}\ifthenelse{\isempty{#2}}{}{_{#2}}}
\NewDocumentCommand{\ten}{ O{} O{} m }{\pmb{\mathscr{#3}}\ifthenelse{\isempty{#1}}{}{^{(#1)}}\ifthenelse{\isempty{#2}}{}{_{#2}}}

\setcopyright{acmlicensed}
\copyrightyear{2025}
\acmYear{2025}
\acmConference[ICAIL 2025]{The 20th International Conference on Artificial Intelligence and Law}{June 16--20,
  2024}{Chicago, IL, USA}

\begin{document}

\title{\texorpdfstring{Bridging Legal Knowledge and AI:\\Retrieval-Augmented Generation with Vector Stores, Knowledge Graphs, and Hierarchical Non-negative Matrix Factorization}{Bridging Legal Knowledge and AI: Retrieval-Augmented Generation with Vector Stores, Knowledge Graphs, and Hierarchical Non-negative Matrix Factorization}}

\author{Ryan C. Barron}
\email{barron@lanl.gov}
\affiliation{
    \institution{Theoretical Division, \\ Los Alamos National Laboratory}
    \city{Los Alamos}
    \state{New Mexico}
    \country{USA}
}
\affiliation{
    \institution{CSEE, UMBC}
    \city{Baltimore}
    \state{Maryland}
    \country{USA}
}
\author{Maksim E. Eren}
\email{maksim@lanl.gov}
\affiliation{
    \institution{Information Systems and Modeling, Los Alamos National Laboratory}
    \city{Los Alamos}
    \state{New Mexico}
    \country{USA}
}
\author{Olga M. Serafimova}
\email{OMSerafimova@hollandhart.com}
\affiliation{
    \institution{Holland \& Hart LLP}
    \city{Santa Fe}
    \state{New Mexico}
    \country{USA}
}
\author{Cynthia Matuszek}
\email{cmat@umbc.edu}
\affiliation{
    \institution{CSEE, UMBC}
    \city{Baltimore}
    \state{Maryland}
    \country{USA}
}
\author{Boian S. Alexandrov}
\email{boian@lanl.gov}
\affiliation{
    \institution{Theoretical Division, \\Los Alamos National Laboratory}
    \city{Los Alamos}
    \state{New Mexico}
    \country{USA}
}

\renewcommand{\shortauthors}{Barron et al.}

\begin{abstract}
Agentic Generative AI, powered by Large Language Models (LLMs) and enhanced with Retrieval-Augmented Generation (RAG), Knowledge Graphs (KGs), and Vector Stores (VSs), represents a transformative technology applicable across specialized domains such as legal systems, research, recommender systems, cybersecurity, and global security, including proliferation research. This technology excels at inferring relationships within vast unstructured or semi-structured datasets.
The legal domain we focus on here comprises inherently complex data characterized by extensive, interrelated, and semi-structured knowledge systems with complex relations. It comprises constitutions, statutes, regulations, and case law. Extracting insights and navigating the intricate networks of legal documents and their relations is crucial for effective legal research and decision-making. Here, we introduce a generative AI system, a jurisdiction-specific legal information retrieval that integrates RAG, VS, and KG, constructed via  Hierarchical  Non-Negative Matrix Factorization (HNMFk), to enhance information retrieval and AI reasoning and minimize hallucinations. In the legal system, these technologies empower AI agents to identify and analyze complex connections among cases, statutes, and legal precedents, uncovering hidden relationships and predicting legal trends—challenging tasks essential for ensuring justice and improving operational efficiency. Our system employs web scraping techniques to systematically collect legal texts, such as statutes, constitutional provisions, and case law, from publicly accessible platforms like Justia. 
It bridges the gap between traditional keyword-based searches and contextual understanding by leveraging advanced semantic representations, hierarchical relationships, and latent topic discovery.
This approach is demonstrated in legal document clustering, summarization, and cross-referencing tasks. The framework marks a significant step toward augmenting legal research with scalable, interpretable, and accurate retrieval methods for semi-structured data, advancing the intersection of computational law and artificial intelligence.

\end{abstract}

\ccsdesc[500]{Computing methodologies~Natural language processing}
\ccsdesc[500]{Applied computing~Law}
\ccsdesc[300]{Computing methodologies~Reasoning about belief and knowledge}
\ccsdesc[300]{Computing methodologies~Ontology engineering}
\ccsdesc[500]{Information systems~Document topic models}
\ccsdesc[500]{Information systems~Content analysis and feature selection}
\ccsdesc[300]{Information systems~Ontologies}
\ccsdesc[300]{Information systems~Document collection models}

\keywords{law, legal knowledge,  nmf, topic labeling, llm, chain of thought, prompt tuning, information retrieval}


\maketitle

\section{Introduction}
\label{sec:introduction}
\input{sections/01_sec_introduction}

\section{Relevant Work}
\label{sec:relevant_work}
\input{sections/02_sec_relevant_work}

\section{Methods}
\label{sec:methods}
\input{sections/03_sec_methods}

\section{Results}
\label{sec:results}
\input{sections/04_sec_results}

\section{Conclusion}
\label{sec:conclusion}
\input{sections/05_sec_conclusion}

\begin{acks}
This research was funded by the U.S. Department of Energy National Nuclear Security Administration’s Office of Defense Nuclear Nonproliferation Research and Development (DNN R\&D), supported by the U.S. DOE NNSA under Contract No. \\89233218CNA000001, as well as by the LANL Institutional Computing Program.
\end{acks}

\bibliographystyle{ACM-Reference-Format}
\bibliography{references.bib}
\end{document}

%% file: sections/01_sec_introduction.tex
The legal domain is uniquely complex, encompassing constitutions, statutes, court rules, regulations, ordinances, and case law. Each source follows distinct structural logics, while constitutions and statutes typically feature hierarchical formatting, case law often appears as long-form unstructured prose. As legal documents are continuously produced, legal professionals and researchers increasingly require computational tools that can go beyond basic keyword-based searches and deliver contextually meaningful, explainable insights. However, traditional legal information retrieval approaches, such as Boolean logic~\cite{westermann2021computer} and lexical matching techniques like TF-IDF~\cite{bithel2021unsupervised}, fall short by missing subtle semantic overlap and deeper contextual dependencies for legal reasoning.

RAG has recently emerged as a compelling paradigm integrating traditional information retrieval's strengths with LLMs' generative capabilities. In the legal domain, RAG systems retrieve relevant documents or knowledge fragments and synthesize them into coherent, grounded answers. This framework mitigates LLM-specific limitations such as hallucinations~\cite{ji2023survey, lewis2020retrieval} by anchoring outputs in authoritative source texts, enhancing performance on tasks like statute interpretation, legal summarization, and case retrieval. Yet to fully realize these benefits, such systems must be supported by infrastructures that can represent, retrieve, and reason over vast, heterogeneous legal corpora.

This work introduces a domain-specific RAG system that integrates three core components to meet these challenges:

\begin{itemize}
    \item \textbf{ VS}: Legal texts are embedded into dense semantic vector spaces using pre-trained models (e.g., BERT~\cite{devlin2019bert}, GPT~\cite{radford2018gpt, NEURIPS20201457c0d6}), enabling retrieval based on contextual similarity rather than surface keyword overlap.
    \item \textbf{ KG}: Legal entities—such as cases, statutes, and doctrines—are interlinked through explicit relationships (citations, shared principles), forming a Neo4j-based graph that supports structured navigation and inference over legal concepts.
    \item \textbf{ HNMFk }: To uncover latent semantic structure, we apply hierarchical NMF (HNMFk) to legal corpora, producing interpretable topic clusters that support document classification, trend discovery, and integration into the KG as abstract semantic anchors.
\end{itemize}

\noindent The resulting architecture fuses high-recall semantic search with logical reasoning via structured knowledge, enabling accurate, explainable responses across a range of legal tasks. Our contributions to this work consist of the following:
\begin{itemize}
    \item \textbf{Jurisdiction-Specific legal information retrieval}: \\
    A pipeline tailored to New Mexico’s legal documents: statutes, constitution, and case law, demonstrating state-level legal domain adaptation.
    \item \textbf{Legal RAG framework tailored to New Mexico’s legal documents}: A modular system combining RAG, VS search, KGs, and LLMs for grounded, explainable legal responses.
    \item \textbf{The first knowledge graph of New Mexico’s legal documents with latent topics}: Our HNMFk extracts and classifies the latent topics of New Mexico's legal documents to construct a Knowledge Graph of metadata and citation links for semantic reasoning.

\end{itemize}

%% file: sections/02_sec_relevant_work.tex
This section reviews contributions across RAG domains, semantic search using vector embeddings, knowledge graph construction, NMF, and legal information systems.

\subsection{Non-Negative Matrix Factorization for Pattern Discovery}
NMF is a dimensionality reduction technique used to uncover latent patterns in data. \cite{lee2001algorithms} analyzed NMF as an interpretable method for extracting features and topics from large datasets, explicitly highlighting its ability to identify meaningful and non-overlapping components. Building on this work, \cite{hoyer2004nonnegative} introduced sparseness constraints for better interpretability,  improving applications of NMF in real-world scenarios through more focused feature selection. In the legal domain, NMF has been valuable for analyzing complex textual data, such as case law and statutes, and assisting with topic discovery and clustering. For instance, \cite{Budahazy2021AnalysisOL} applied NMF to legal documents to extract latent topics and visualize relationships, demonstrating how NMF’s interpretable structure aids researchers in identifying underlying topics not readily apparent in raw text. They even applied NMF hierarchically to find fine-grained topics. More recently, \cite{Li2022GuidedSN} proposed a guided semi-supervised NMF approach for topic discovery in legal documents, using domain knowledge to steer factorization and ensure the extracted topics remain highly relevant. This semi-supervised extension bridges the gap between fully automated unsupervised techniques and expert-driven analysis.  NMF’s use in legal contexts is significant, particularly for interpretable results on large textual datasets.

\subsection{Retrieval-Augmented Generation}
RAG has emerged as a foundational approach for improving AI systems across various domains, including law. \citet{lewis2020retrieval} introduced a framework that dynamically retrieves relevant documents to ground generative outputs, achieving notable gains in accuracy. Building on this idea, \citet{guu2020realm} proposed a retrieval-augmented pretraining method that integrates external knowledge for improved downstream task performance, while \citet{izacard2021leveraging} demonstrated the effectiveness of retrieval in open-domain question answering. These advances lay the groundwork for applying RAG to the legal sector, where the method’s ability to ground LLMs in authoritative texts reduces hallucinations and increases accuracy in tasks including law retrieval, statutory reasoning, and judgment prediction. Notable examples include CBR-RAG, which incorporates Case-Based Reasoning to structure retrieval for legal QA \cite{10.1007/978-3-031-63646-2_29}, and LegalBench-RAG. This benchmark suite tailors evaluation metrics to the demands of legal information synthesis \cite{pipitone2024legalbenchragbenchmarkretrievalaugmentedgeneration}. Parallel work has demonstrated RAG’s capabilities in other domains, such as malware data analysis, by combining embeddings, KGs, and NMF \cite{barron2024domainspecificretrievalaugmentedgenerationusing}. Other works show how LLMs can dynamically decide when and what to retrieve to improve legal reasoning \cite{asai2023retrieve} and how multi-step legal judgment prediction can benefit from iterative retrieval and generation \cite{cui2023multistep}, further demonstrating the effectiveness of combining retrieval strategies with LLMs.

\subsection{Semantic Search with Vector Embeddings}
Semantic search operates on dense vectors to find the deeper semantic relationships in texts, going beyond keyword matching to proper context retrieval \cite{karpukhin2020dense, guu2020realm}. This search is especially valuable in the legal domain, where queries often demand conceptual rather than surface-level understanding. Domain-specific pretraining has become increasingly important, as in LEGAL-BERT \cite{chalkidis2020legalbert}, outperforming general-purpose models by better capturing legal language nuances. Benchmarks like LeCaRD \cite{ma2021lecard} show the effectiveness of dense retrievers, such as SBERT, in legal case retrieval of Chinese law, and the abilities of hybrid approaches that integrate lexical and dense retrieval methods. Hierarchical transformer architectures \cite{xiao2021hierarchical} and long-context models like Longformer \cite{beltagy2020longformer} further address the challenges of lengthy legal documents. In contrast, citation-driven approaches such as SPECTER \cite{cohan2020specter} leverage metadata to improve retrieval. Furthermore, work on neighborhood contrastive learning for scientific document representations (SciNCL) \cite{ostendorff2022neighborhood} demonstrates how controlled nearest neighbor sampling over citation graph embeddings can provide continuous similarity. This strategy could also inform citation-based retrieval improvements in legal domains. Challenges remain for scaling, explainability, and heterogeneous data sources in real-world legal workflows.

\subsection{Knowledge Graphs in Legal Research}
Although the legal knowledge is jurisdiction-specific, various KGs have been widely adopted across various legal systems in legal research, offering structured representations that support case retrieval, citation analysis, and question answering. In the United States, researchers already have used KGs for legal knowledge extraction \cite{Sovrano2020LegalKE}, researchers also built various domain-specific graphs to model regulatory frameworks \cite{Dong2021AKG}, and explored their potential in addressing concerns of trust, privacy, and transparency in legal AI systems \cite{a9b58e3cf6ea4306b01b44386cbf9c1f}. 

In other jurisdictions, similar KG methodologies have proven valuable. In India, KGs have been applied to recommend comparable legal cases \cite{dhani2024similarcasesrecommendationusing}. For example, in China, they have been used to link statutes and case law to improve retrieval and to enhance case law search in Chinese courts \cite{zhou2024automaticknowledgegraphconstruction, BI2022109046}. In Germany, Milz et al.\ conducted a structural analysis of a legal citation network using KGs to map citation patterns in German court decisions \cite{German_Legal_Citation}.

Beyond case retrieval, KGs have also been incorporated into knowledge-aware machine reading systems designed for legal question answering \cite{long2021kalm}. These international efforts highlight the versatility of KGs in capturing the intricacies of legal systems. In this work, we build on these foundations within the context of New Mexico statutes and case law, demonstrating the applicability of KGs to state-level U.S. legal research and retrieval.

\subsection{Legal Systems and Case Retrieval}
Legal information systems have evolved rapidly with the advent of neural architectures and hybrid retrieval pipelines, enabling precise tasks such as precedent retrieval, statute matching, and judgment prediction \cite{chalkidis2019neural, robaldo2020neural, chalkidis2021paragraph}. Benchmarks like LeCaRD \cite{ma2021lecard} and LexGLUE \cite{chalkidis2022lexglue}, along with LEGAL-BERT-based systems \cite{chalkidis2020legalbert}, have demonstrated the capability of these neural methods to improve accuracy in analyzing legal corpora. In particular, researchers have leveraged structured reasoning with transformers and graph representations to link statutes and precedents, as evidenced in the COLIEE competition \cite{rabelo2021coliee}. Nonetheless, several limitations remain: data scarcity and jurisdictional bias continue to restrict the generalizability of such models. At the same time, resource-intensive retrievers like BERT-based cross-encoders \cite{shao2022bert} have challenges scaling to large-scale legal databases. Earlier works in juris-informatics have highlighted the potential of automating legal reasoning and document analysis \cite{ashley2017artificial}, laying the groundwork for modern approaches that fuse knowledge graphs, transformers, vector stores, and agent-oriented RAG pipelines to deliver more explainable and efficient legal workflows.

%% file: sections/03_sec_methods.tex
Laws and their interpretations are limited in application to their respective jurisdictions. This examines the State of New Mexico, using the available Supreme Court and Court of Appeals case law, the state constitution, and state statutes as the primary data resource.
\subsection{Overview and Interconnect}
\begin{figure}[h!]
\vspace{-0.5em}
\centering
    \includegraphics[width=.4\linewidth]{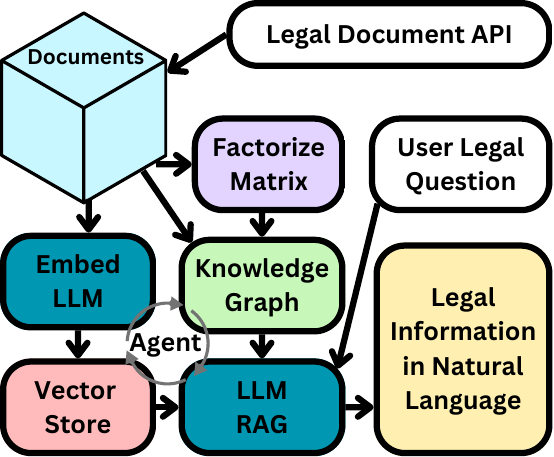}
\vspace{-.6em}
    
    \caption{System Overview}
    \Description{Diagram showing the system components in the legal AI framework.}
    \label{fig:law_overview}
\hfill
\vspace{-2em}
\end{figure}

Our jurisdiction-specific AI system integrates RAG, VS, KG, HNMFk, and LLMs in a modular, agentic architecture for New Mexico's legal information retrieval and reasoning, shown in Figure \ref{fig:law_overview}.

Queries enter the system through a natural language interface. The input is embedded and used to perform two operations: (1) a semantic retrieval against a vector database containing constitutional provisions, statutes, and segmented case law, and (2) a traversal of the Neo4j-based legal knowledge graph containing latent HNMFk topics, citations, and structural metadata from the legal corpus.

An orchestration layer evaluates both results, aligning retrieved documents and KG nodes with the user's query context. The agent performs reasoning across case-to-case, statute-to-case, or doctrine-level connections if a relevant KG subgraph is identified. Otherwise, it defaults to the highest-confidence passages from vector search.

The selected materials are passed to an LLM agent in both cases, where a final response is generated, grounded in precise citations. If the user requests clarification or follow-up, the system maintains a short-term state and re-evaluates context via VS and KG pathways.

\subsection{Data Collection}
To compile a research-oriented corpus of legal documents, we used Justia’s publicly accessible resources in full compliance with their terms of service and robots.txt guidelines. A restricted, responsible crawler was developed to access only permitted content such as statutes, constitutional provisions, and public case law, while respecting rate limits and avoiding disallowed sections. The scraper incorporates request delays, monitors HTTP status codes, and applies exponential backoff to ensure ethical data collection.

Starting from Justia’s landing pages, the crawler identifies valid navigation links, resolves absolute URLs, and recursively follows permitted content. It extracts metadata and full text for statutes, constitutional provisions, citations, and judicial opinions for New Mexico’s Supreme Court and Court of Appeals.

Compliance and transparency are maintained through:
\begin{itemize}
    \item \textbf{Logging} to track URLs, prevent duplication, and resume interrupted crawls.
    \item \textbf{Rate limiting} to avoid overloading Justia’s servers.
    \item \textbf{Respect for technical restrictions} in robots.txt / TOS.
    \item \textbf{Strictly non-commercial use}, limited to scholarly research.
\end{itemize}

These measures preserve ethical and legally compliant data collection that is aligned with Justia’s content usage policies.

\subsection{Dimension Reduction in Legal Texts}
Legal documents- constitutions, statutes, and case law- are traditionally organized into chapters, articles, and sections, but these structures do not always match the latent patterns revealed through factorization. Each document type is analyzed separately using non-negative tensor and matrix factorization to explore these hidden relationships. A TF-IDF matrix, \( \mathbf{X} \), is first constructed from the cleaned corpus. Constitutional provisions, statutory clauses, and case law paragraphs form the units of analysis for clustering.

In this study, we use \textbf{Tensor Extraction of Latent Features} (\textbf{T-ELF})\footnote{\url{https://github.com/lanl/T-ELF}} \cite{TELF}, combined with automatic model selection, to decompose \( \mathbf{X} \) into coherent topic H-clusters. \textbf{T-ELF} efficiently identifies latent topics, grouping constitutional provisions around themes such as “separation of powers” and clustering statutes and case law based on regulatory objectives and recurring legal doctrines, respectively. The optimal number of clusters, \( k \), is determined using silhouette scores and is accelerated 
through a binary search \cite{barron2024binarybleedfastdistributed}.

\subsubsection{Application of Non-Negative Matrix Factorization to Legal Texts}

NMF approximates the matrix \( \mathbf{X} \) with two non-negative matrices, \( \mathbf{W} \) and \( \mathbf{H} \) such that $\mathbf{X} \approx \mathbf{W} \cdot \mathbf{H}$, where \( \mathbf{W} \) describes how terms distribute over topics, and \( \mathbf{H} \) describes how these topics distribute across documents. Constitutional articles and sections reveal underlying governance or civil rights themes; statute clauses highlight regulatory objectives, and segmented judicial opinions expose common doctrines and legal reasoning patterns.

\subsubsection{Automatic Model Determination Using NMFk}

A central challenge in applying NMF is selecting the best number of latent features (\( k \)). We use \textit{NMFk}~\cite{alexandrov2020patent}, which combines clustering stability with reconstruction accuracy. Bootstrap resampling generates slightly perturbed versions of the original matrix, and repeated decompositions measure how consistently clusters form. Silhouette scores help ensure cohesive, well-separated topics, while reconstruction error verifies that the model effectively captures patterns in the original data.

By adapting this NMF approach to a hierarchical setting, legal texts can be organized into a tree-like structure. Constitutions may be segmented into articles and sections, statutes into chapters and clauses, and case law into layered precedents and sub-issues. This hierarchical perspective enhances the discovery of latent relationships at multiple levels of granularity, facilitating deeper analyses of large-scale legal corpora.

\subsection{Knowledge Graph}
Features derived from \textbf{T-ELF} and document metadata are transformed into a series of head, entity, and tail relations, forming directional triplets integrated into a Neo4j KG \cite{neo4j2023}.

In the legal context, the KG incorporates metadata and latent features extracted from constitutions, statutes, and case law. The primary nodes in the graph represent legal documents, including constitutional provisions, statutory sections, and judicial opinions. These nodes are enriched with metadata such as titles, hierarchical identifiers (e.g., chapter and section numbers), jurisdiction, court names, decision dates, and topics derived from latent features. 

\begin{figure}[h]
    \vspace{-1em}
    \centering
    \includegraphics[width=.7\columnwidth]{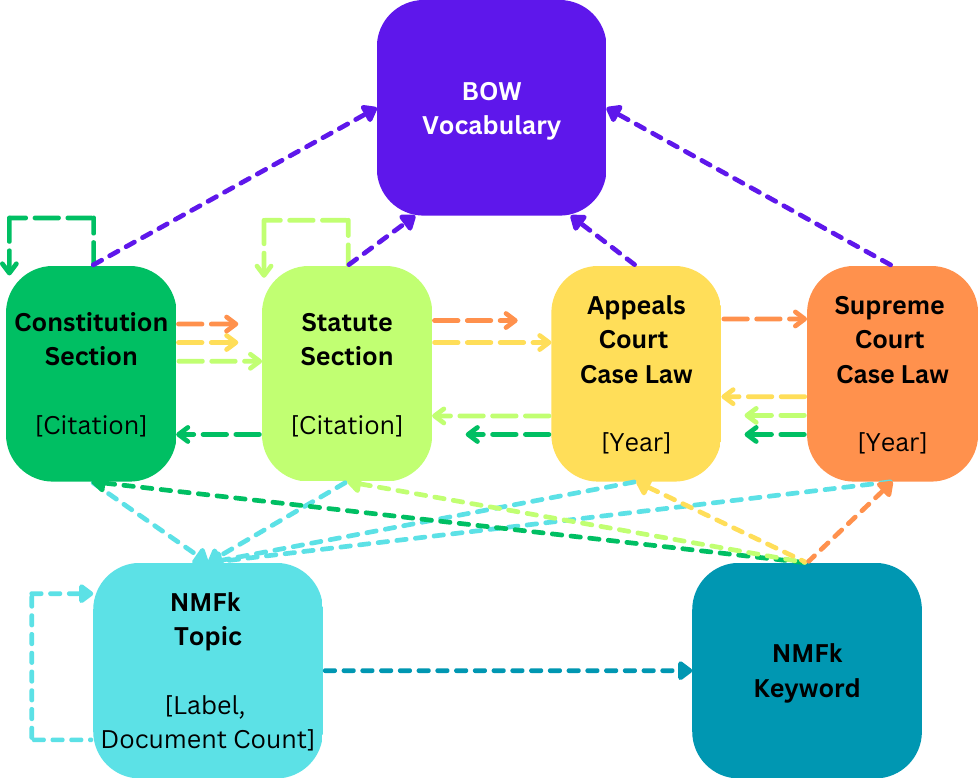}
    \vspace{-1em}
    \caption{Knowledge Graph Schema, with the primary identifier in bold and attributes in brackets.}
    \Description{Diagram showing a knowledge graph schema where primary nodes are bolded and attributes are shown in brackets.}
    \label{fig:law_schema}
\end{figure}
\vspace{-1em}

Edges in the KG establish relationships between nodes to represent the interconnected nature of legal documents. For instance:

\begin{itemize}
    \item \textbf{Constitutional Nodes:} Linked to statutory provisions or judicial interpretations that reference or rely on specific constitutional clauses.
    \item \textbf{Statutory Nodes:} Connected to cases interpreting the statute or related provisions within the same legislative framework.
    \item  \textbf{Case Law Nodes:} Interlinked based on shared topics, common legal principles, or hierarchical relationships in appellate decisions.
\end{itemize}

This graph structure enables the RAG system to query and retrieve legal documents based on semantic similarity and explicit relationships. For example, a query about "due process" might retrieve the relevant constitutional clause, cases that discuss its interpretation, and statutory provisions impacted by those rulings. By combining metadata and latent features, the KG supports advanced reasoning and logical traversal, enhancing the precision and depth of legal information retrieval.

The Bag-of-Words (BOW) vocabulary, i.e., the unique unigrams of the corpus, was generated through standard NLP processing, measuring their frequency of tokens in documents. The most frequent 50 unigrams in a topic, rows of $\mathbf{W}$, define the top keywords in this topic. This included stop word removal, special character removal, etc. %

\subsection{Vector Store}
A vector database was implemented to manage and index legal documents, improving the RAG process for legal research. Using Milvus~\cite{2021milvus}, the system stores vectorized representations of constitutions, statutes, and case law, treating each document type uniquely to ensure contextually relevant retrieval.

Constitutional provisions are split into paragraphs, each with a unique ID, and vectorized using OpenAI’s text-embedding-ada-002~\cite{openai_api} model for granular semantic search. Statutes are divided into sections or clauses, with metadata like chapter and section titles added for precise retrieval. Case law, being unstructured, is chunked into logical units, preserving flow and indexed with metadata such as case name and citation.

Based on the query's focus, the RAG application queries the database to retrieve relevant fragments—constitutional paragraphs, statutory clauses, or case law sections. Retrieved text is synthesized into responses, allowing the LLM to cite specific paragraphs or clauses, ensuring traceability and accuracy.
The system leverages a connected knowledge graph to explore related amendments, judicial interpretations, or precedent cases for additional context. This integration provides a robust, comprehensive retrieval process tailored to the complexities of legal research.

\subsection{Evaluation Metrics}
\label{eval_metrics_description}
In evaluating AI-generated responses against reference texts, metrics assess content quality, semantic similarity, and factual consistency. ROUGE-L (Recall-Oriented Understudy for Gisting Evaluation) evaluates lexical overlap of the longest common subsequence to assess structural similarity \cite{lin2004rouge}.  Natural Language Inference (NLI) entailment quantifies whether the generated text is logically entailed by the reference, based on pretrained models like BART \cite{williams2018broad}. SummaC evaluates factual consistency by aggregating entailment scores across sentence pairs between the hypothesis and source \cite{laban2022summac}. FactCC is another entailment-based model that identifies factual inconsistencies in summarization outputs by fine-tuning on labeled correctness data \cite{kryscinski2020evaluating}. Additionally, we leverage human evaluation on responses for a small subset of the questions.

\begin{figure}[ht]
    \vspace{-.9em}
    \centering
    \includegraphics[width=.8\columnwidth]{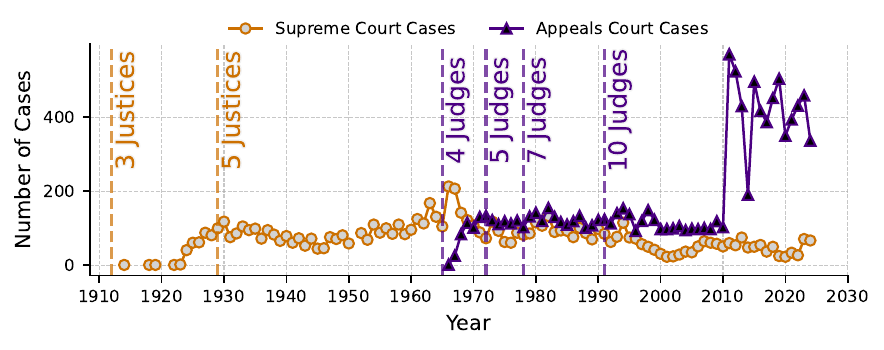}
    \vspace{-1.5em}
    \caption{New Mexico Supreme/Appeals case counts per year.}
    \Description{Chart showing the number of New Mexico Supreme Court and Appeals Court cases for each year.}
    \vspace{-1.5em}
    \label{fig:court_cases}
\end{figure}

%% file: sections/04_sec_results.tex
This section presents the resulting legal texts and their hierarchical decomposition, offering a detailed breakdown of the sections and cases within each document type. The following results illustrate the scope and depth of the collected data, providing a foundation for further exploration of trends and patterns across the legal corpus.

\subsection{Dataset}
After collecting and structuring the data, the legal documents were organized into 265 Constitutional provisions,28,251 Statutory sections, and Case Law comprising 10,072 Court of Appeals cases and 5,727 Supreme Court cases. Figure~\ref{fig:court_cases} shows the trends in these Supreme Court and Court of Appeals cases over the years, as available from Justia \cite{justia}, which also includes the expansions and creations of the courts themselves.

\begin{figure*}[ht]
    \centering
    \begin{subfigure}[t]{0.24\textwidth}
        \centering
        \includegraphics[width=\textwidth]{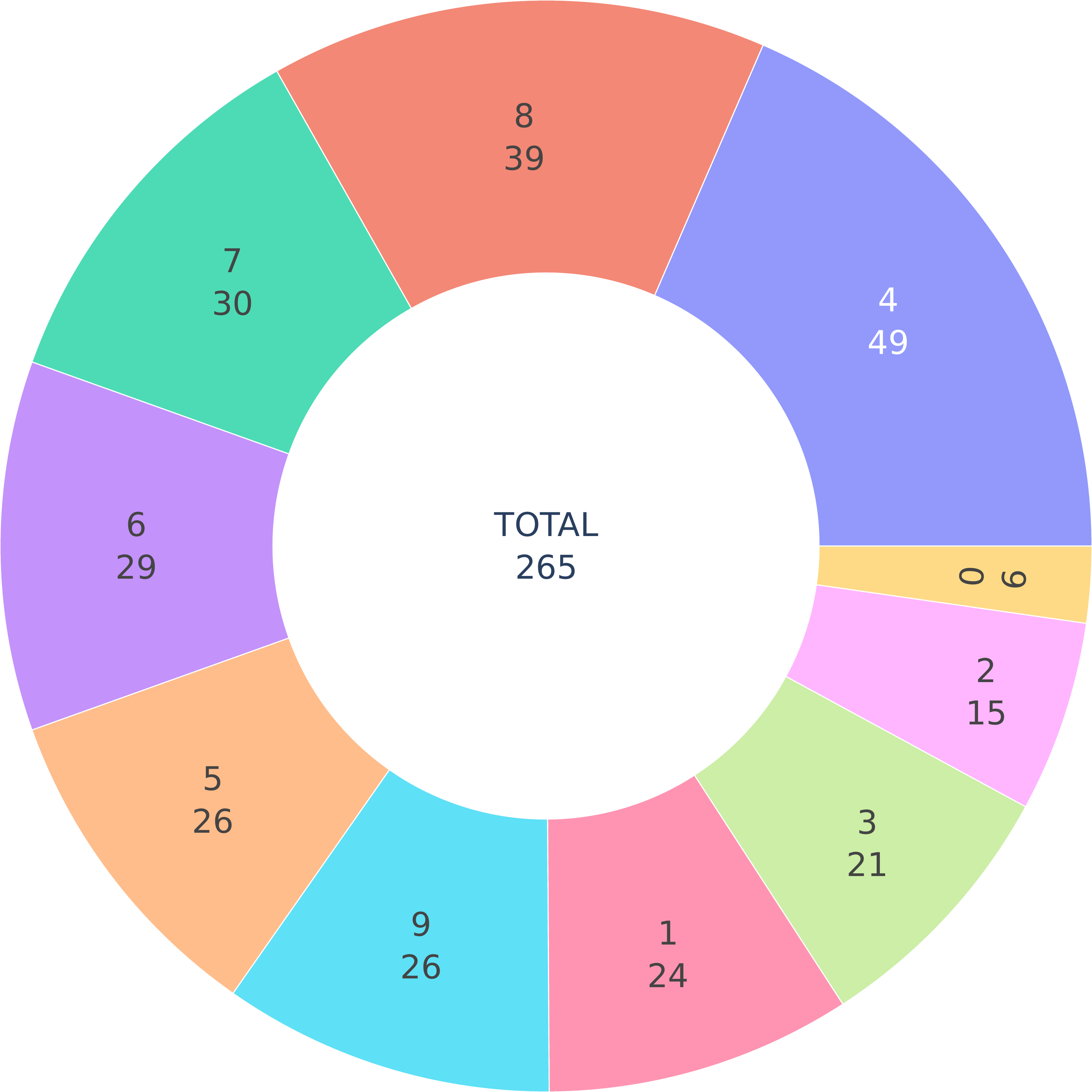}
        \caption{10 latent topics from \\ \textbf{Constitutional Provisions}}
        \label{fig:decomp_const}
    \end{subfigure}
    \hfill
    \begin{subfigure}[t]{0.24\textwidth}
        \centering
        \includegraphics[width=\textwidth]{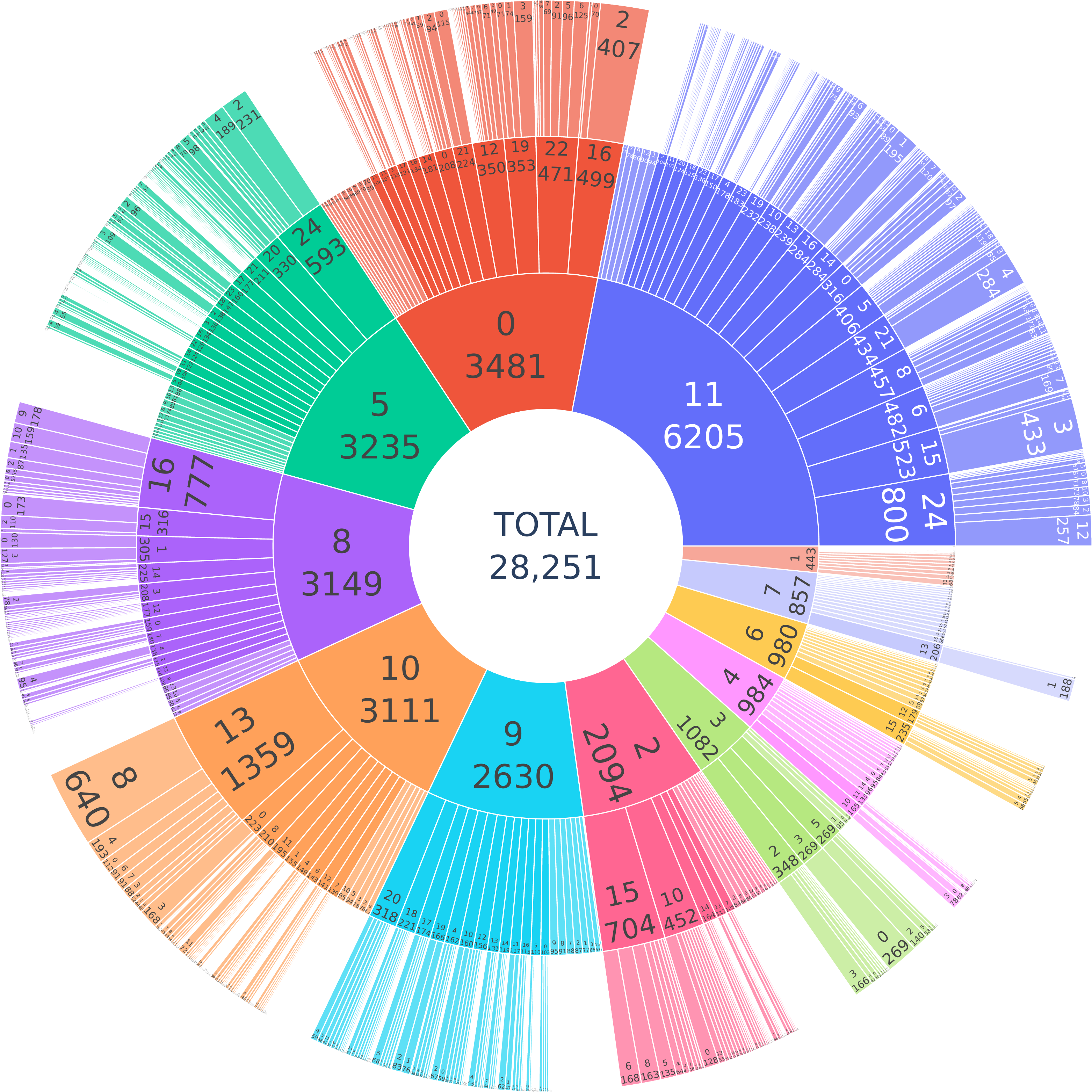}
        \caption{985 latent topics from \\ \textbf{Statutory Sections}}
        \label{fig:decomp_stat}
    \end{subfigure}
    \hfill
    \begin{subfigure}[t]{0.24\textwidth}
        \centering
        \includegraphics[width=\textwidth]{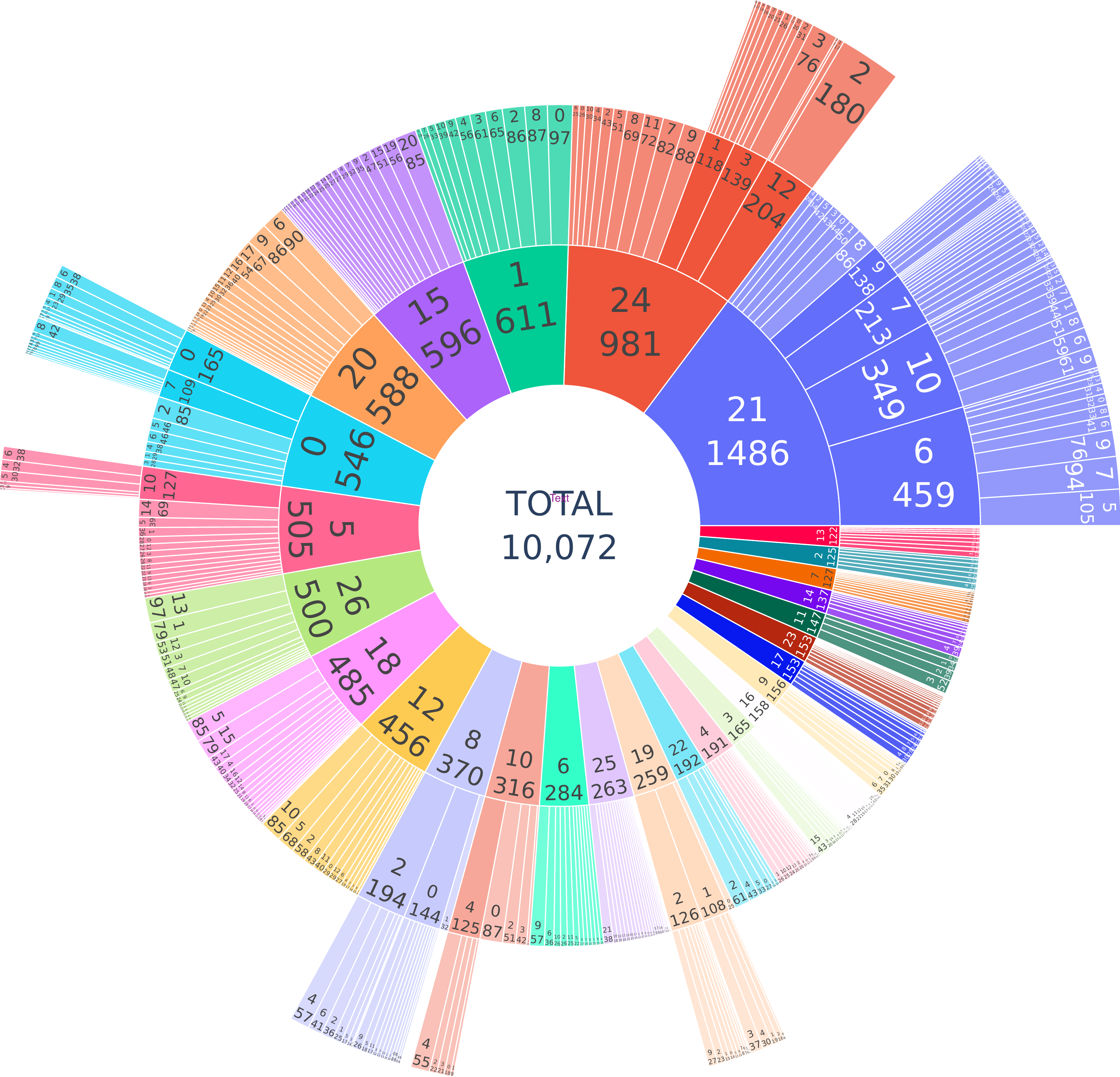}
        \caption{420 latent topics from \\ \textbf{Court of Appeals Cases}}
        \label{fig:decomp_sup}
    \end{subfigure}
    \hfill
    \begin{subfigure}[t]{0.24\textwidth}
        \centering
        \includegraphics[width=\textwidth]{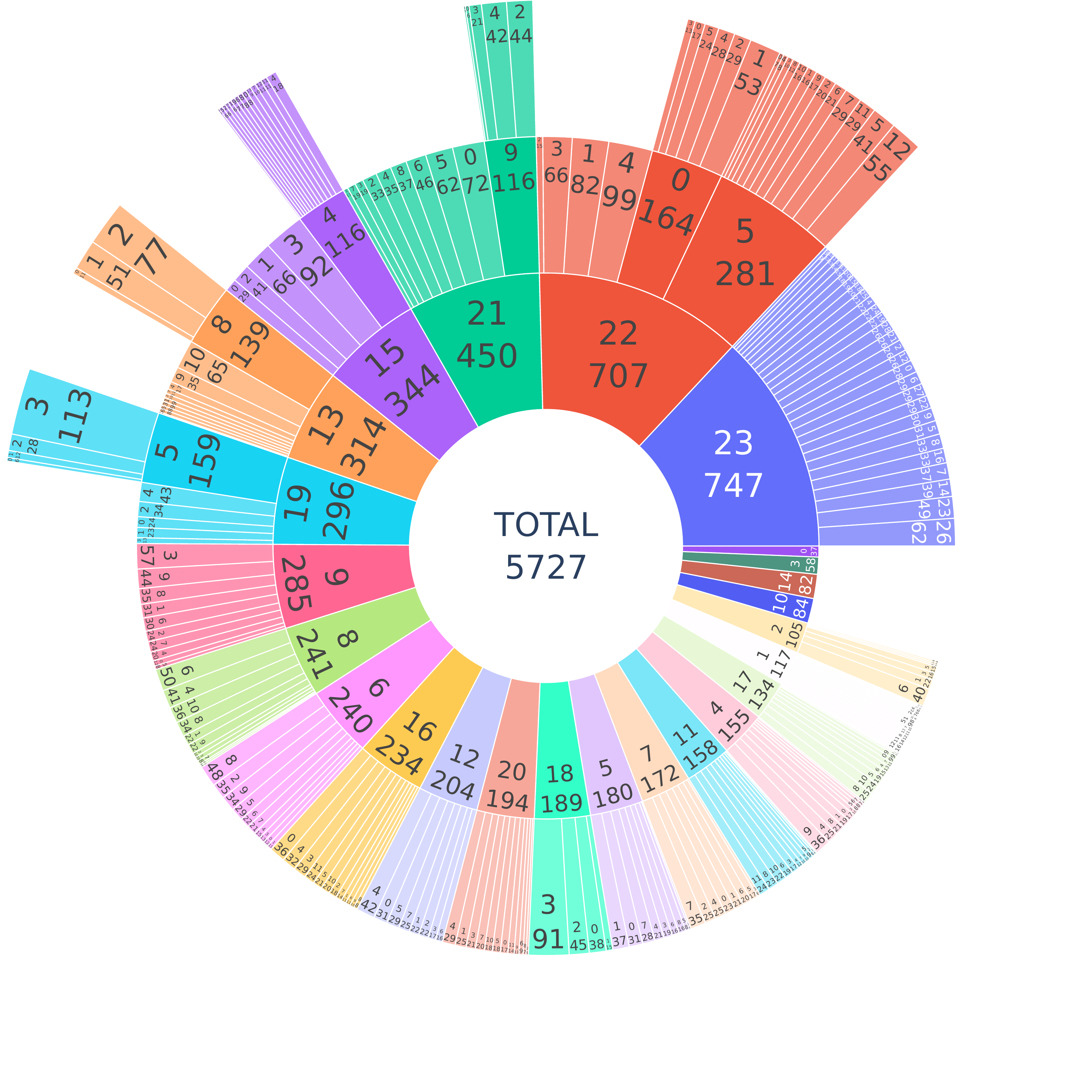}
        \caption{248 latent leaf topics from \textbf{Supreme Court Cases}}
        \label{fig:decomp_app}
    \end{subfigure}
    \vspace{-.5em}
    \caption{Legal Documents from New Mexico hierarchically decomposed. The Constitution only had enough documents to decompose the first depths, whereas the other three sources continued to the terminal depth of 2 (a hyper-parameter of decomposition). Each H-cluster has a natural language label, where depth-0 from each can be seen in Tables \ref{tab:Constitution}, \ref{tab:statutes}, \ref{tab:supreme}, and \ref{tab:appeals}. }
    \Description{Four sunburst charts, each showing hierarchical topic decomposition of New Mexico legal documents. The Constitution chart has fewer layers; statutes, court of appeals, and supreme court cases go deeper. Each segment represents a latent topic.}
    \label{fig:decomp_all}
    \vspace{-1.5em}
\end{figure*}

\subsection{Decomposition}
The four component data types were decomposed hierarchically with NMFk. The vocabulary for building the TF-IDF matrix was collected using specific parameters for each part. For the Constitution vocabulary, the minimum token document frequency (DF) was set to 5 documents, with a maximum token DF of 80\% of the corpus, resulting in a final size of 416 tokens. For the Statutes vocabulary, the minimum token DF was set to 30 documents, with a maximum token DF of 70\% of the corpus, yielding 7,508 tokens. The Court of Appeals vocabulary used a minimum token DF of 50 documents (cases) and a maximum token DF of 70\% of the corpus, resulting in 10,189 tokens. Last, the Supreme vocabulary employed a minimum token DF of 50 papers and a maximum token DF of 70\% of the corpus, with a final size of 8,425 tokens.
The maximum depth was set to 2, of which only the constitution sections did not reach due to the limited number of sections. The limiting factor of further decompositions was 100 documents, so if the preceding cluster had $\geq 100$, the cluster would decompose and stop otherwise. As seen in Figure \ref{fig:decomp_const}, the largest H-cluster is from cluster 4 in dark blue with 49 sections, with the fewest documents in yellow with six sections in H-cluster 0. The other three decompositions can be examined in the larger Figure \ref{fig:decomp_all}, where the court cases, there are 10 leaf H-clusters in the constitution decomposition, 985 leaf H-clusters in the statutes decomposition,  420 leaf H-clusters in the Court of Appeals cases, and 248 leaf H-clusters in the Supreme Court cases. From the methods in \cite{topic_tag}, each H-cluster throughout the decomposition hierarchies has LLM-generated labels for ease of reference and quick insight. Labels for the first decomposition depth for the constitution can be observed in Table \ref{tab:Constitution},  the statutes in Table \ref{tab:statutes}, the Supreme court in Table \ref{tab:supreme}, and the appeals in Table \ref{tab:appeals}. The depth-0 H-clustering corresponds to the first ring radial from the center totals in Figure \ref{fig:decomp_all}.

\definecolor{darkgreen}{rgb}{0.0, 0.6 , 0.0}
\definecolor{lightgreen}{rgb}{0.0, 0.9, 0.0}
\definecolor{goldyellow}{rgb}{1.0, 0.84, 0.0}
\definecolor{darkorange}{rgb}{1.0, 0.55, 0.0}
\definecolor{lightblue}{rgb}{0.0, 0.9, 1.0}
\definecolor{darkpurple}{rgb}{0.4, 0.0, 0.4}

\begin{table}[!htbp]
    \tiny  
    \caption{NM Constitutional Depth-0 H-Clusters}
    \vspace{-1em}
    \label{tab:Constitution}
    \renewcommand{\arraystretch}{0.9} 
    \begin{tabular}{@{}p{0.02\columnwidth}p{0.89\columnwidth}@{}}
        \textbf{\#} & \textbf{Label} \\
        \toprule
        0 & Irrigation and Water Resource Management Principles \\
        1 & Regulation of Private Sector Influence on Public Schools and Education Services \\
        2 & Branches of Government Structure and Functionality \\
        3 & Education Funding for New Mexico's Educational Institutions \\
        4 & Legislative Proceedings and Lawmaking Activities Enacted During Sessions \\
        5 & Territorial Transition: Constitutional Ratification and Statehood Provisions\\
        6 & Governance, Land, and Taxation Framework \\
        7 & Municipal and County Financial Obligations and Liabilities Management \\
        8 & Judicial Power and Jurisdictional Frameworks \\
        9 & Public Service Election Governance Structure and Processes \\
        \bottomrule
    \end{tabular}
\end{table}

\begin{table}[!htbp]
    \tiny 
    \caption{NM Statutory Depth-0 H-Clusters}
    \vspace{-1em}
    \label{tab:statutes}
    \renewcommand{\arraystretch}{0.9}  
    \begin{tabular}{@{}p{0.02\columnwidth}p{0.89\columnwidth}@{}}
        \textbf{\#} & \textbf{Label} \\
        \toprule
        0  & Municipal Court Civil Cases Involving Children's Rights \\
        1  & Public Education Infrastructure Management Systems \\
        2 &  Criminal Codes, Local Governance Boundaries, Licensing Rules\\
        3  & Comprehensive Emergency Health and Human Services Response Framework \\
        4  & Taxation and Revenue Collection Oversight \\
        5 & Military Decorations, Licensing Procedures, Governance Boards \\
        6  & Regulatory Insurance Contract Law and Policy Analysis \\
        7  & Public Obligations Financing and Project Bonds Issuance \\
        8  & State Government Investment Grants for Education and Economic Development \\
        9  & Irrigation and Water Rights Regulations \\
        10 & Military Honors, Discrimination Penalties, and Trust Authority \\
        11 & Corporation Governance Framework and Regulatory Compliance \\
        \bottomrule
    \end{tabular}
\end{table}

\begin{table}[!htbp]
    \tiny  
    \caption{NM Appeals Court Case Law Depth-0 H-Clusters}
    \vspace{-1em}
    \label{tab:appeals}
    \renewcommand{\arraystretch}{0.9}  
    \begin{tabular}{@{}p{0.02\columnwidth}p{0.89\columnwidth}@{}}
        \textbf{\#} & \textbf{Label} \\
        \toprule
        0  & Real Property Rights and Interests \\
        1  & Civil Liability and Injury Issues Arising from Healthcare Services \\
        2  & Parental Rights and Custody Proceedings Involving Disputed Parental Fitness \\
        3  & Motor Vehicle Insurance and Liability Claims Processing Procedures \\
        4  & Mortgage Foreclosure and Secured Lending Frameworks \\
        5 & Arraignments, Tribal Jurisdiction, Divorce, Bond Conditions, Motor Licensing \\ 
        6  & Courtroom Advocacy and Representation Strategies \\
        7  & Petitioner's Right to Parental Decision Making \\
        8  & Fourth Amendment Protections Against Unlawful Searches and Seizures \\
        9  & Taxation of Gross Receipts and Sales Transactions in a Business Context \\
        10 & Key Legal Concepts Related to Sexual Offenses \\
        11 & Child Protective Services Laws and Regulations \\
        12 & Work-Related Injury and Disability Compensation Process \\
        13 & Healthcare Contract Dispute Resolution Processes \\
        14 & Marital Property and Financial Disposition \\
        15 & Legal Proceedings Involving Jury Determination of Guilt \\
        16 & Probation Supervision and Monitoring Oversight Process \\
        17 & Juvenile Protection, Miranda Rights, Evidence Collection \\
        18 & Firearms, Substances, DWI, and Sentencing \\
        19 & Sentencing Enhancements for Serious Repeat Offenders \\
        20 & Traffic Stop Under Suspicion with Mandatory Blood Alcohol Testing \\
        21 & Criminal Proceedings Trial Litigation Documentation and Record-Keeping Procedures \\
        22 & Workers' Rights and Insurance Benefits \\
        23 & Children's Welfare and Family Reunification Efforts \\
        24 & Public Municipal Legal Frameworks and Governance Structure \\
        25 & Business Disputes, Wrongful Injury, Taxation Appeals, Workers’ Compensation \\ 
        26 & Administrative Disputes,  Native American Legal Memorandums \\
        \bottomrule
    \end{tabular}
\end{table}

\begin{table}[!htbp]
    \tiny  
    \caption{NM Supreme Court Case Law Depth-0 H-Clusters}
    \vspace{-1em}
    \label{tab:supreme}
    \renewcommand{\arraystretch}{0.9} 
    \begin{tabular}{@{}p{0.02\columnwidth}p{0.89\columnwidth}@{}}
        \textbf{\#} & \textbf{Label} \\
        \toprule
        0  & Arbitration of Contract Disputes and Judicial Decision-Making in Motor Vehicle Cases \\
        1  & Revenue and Taxation Frameworks in Governance and Administration \\
        2  & Mineral Rights Leases \\
        3  & Native American Self-Governance and Tribal Jurisdictional Frameworks \\
        4  & Municipal Zoning Ordinances and Regulations of Local Governance Areas \\
        5  & Damages Award for Wrongful Conduct Against Business Partners \\
        6 & Jurisdictional Appeals, Juvenile Sentencing, Felony Enhancements, Parole Terms \\
        7  & Post-Divorce Asset Distribution and Management Strategies \\
        8  & Constitutional Challenges to Public Education Governance \\
        9  & Secured Financial Instruments and Property Rights \\
        10 & Water Resource Allocation and Management \\
        11 & Electric Utility Rate Regulation Oversight Authority \\
        12 & Law Enforcement Procedures and Rights Protection under Fourth Amendment Protections \\
        13 & State Legislative Review Process Decisions \\
        14 & Denial of Petition for Habeas Corpus \\ 
        15 & Workers' Compensation Process for Work-Related Injuries and Disabilities \\
        16 & Real Estate Boundary Disputes and Conveyance Matters \\
        17 & Parental Rights and Legal Guardianship Proceedings \\
        18 & Appellant's Argument Against the Ruling of a Trial Court \\
        19 & Document Terms and Conditions Regarding Business Transactions \\
        20 & Mutual Insurance Policies for Vehicle and Individual Coverage \\
        21 & Accident resulting from driver error on public roadways leads to legal consequences \\
        22 & Court proceedings involving criminal trials and testimonial evidence \\
        23 & Civil Litigation Proceedings and Trials \\
        \bottomrule
    \end{tabular}
\end{table}

\subsection{Knowledge Graph}
The four data parts, 265 constitutional provisions, 28,251 Statute sections, 5,727  Supreme Court cases, and 10,072 Court of Appeals cases, were inserted into the neo4j \cite{neo4j2023} knowledge graph. 
The graph's number of nodes and edges can be seen in Table \ref{tab:neo4j_counts}, where edge counts are where the triplet's tail originates with the row's node. The legal citations were collected by iterating the text of each case or section into chat-gpt-3.5-turbo with the following prompt: "You are an expert legal document analyzer. Your job is to find all references to the Constitution, Case Law, or Statutes in the text." The result was that the LLM acted like a named entity extractor, such that any citations in the text were pulled out in an enumerated list. The citations mainly included the cases, statutes, and constitution of New Mexico but also had references to the United States Constitution, New Mexico Administrative Code, and New Mexico Rules Annotated (NMRA). The NMRA had many references to Uniform Jury instructions and the rules of criminal and civil proceedings.

\begin{figure}[ht]
    \centering
    \includegraphics[width=\columnwidth, trim=15 0 15 5, clip]{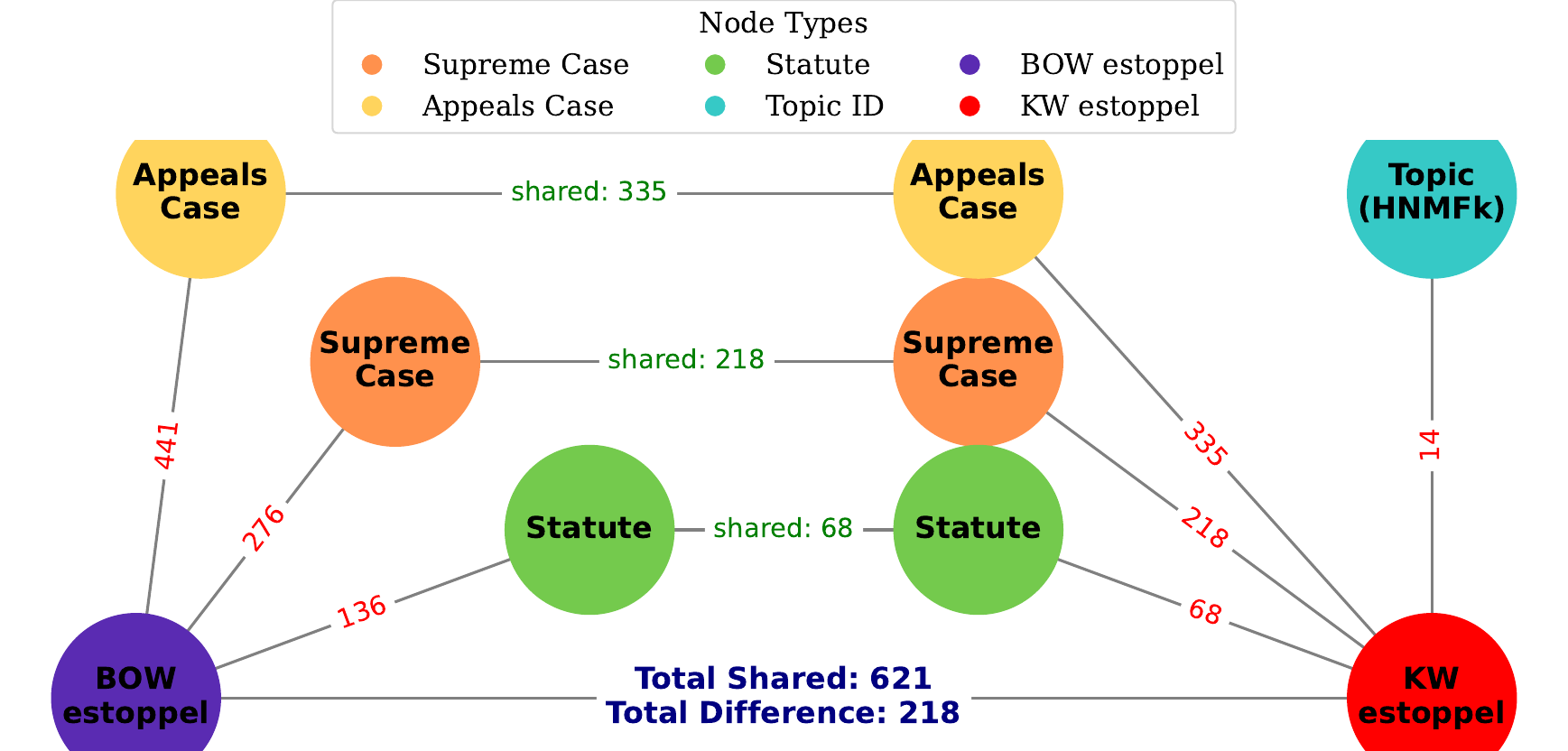}
    \vspace{-2em}
    \caption{ Examination of `Estoppel' relating to being a keyword in topics, vs bag of word vocabulary. }
        \Description{Figure showing how the term 'Estoppel' appears differently when used as a keyword in topics compared to its frequency in the bag-of-words vocabulary.}
    \vspace{-1em}
    \label{fig:estoppel}
\end{figure}

In figure \ref{fig:estoppel}, the NMFk topic keyword and a bag-of-words vocabulary were both queried for `estoppel.' The NMFk keyword is red, the topics show in a light blue node, and the BOW node is purple. Three node types occur for this keyword: the statutes are green nodes, the Court of Appeals cases are yellow, and the Supreme Court cases are orange. The constitution neither clustered over the term nor mentions it, which is not represented in Figure \ref{fig:estoppel}. There is a difference in the data that have topics associated with estoppel vs all of the documents that mention 'estoppel'. Still, not all documents mentioning `estoppel' were clustered with the word, which means that other terms and concepts from those terms had more importance for the documents on the left side of the image than `estoppel'. There are 14 topics, 441 Court of Appeals cases,  276 Supreme Court Cases, and 136 Statutes in Figure \ref{fig:estoppel}. 

\begin{table}[t!]
\centering
\caption{Neo4j Node and Edge Overview}
    \vspace{-1em}
\label{tab:neo4j_counts}
\definecolor{headerblue}{HTML}{CFE2F3} 
\definecolor{lightgray}{gray}{0.92}     
\begin{tabular}{@{}lrrr@{}}
  \toprule
  \textbf{Node Type}      & \textbf{Nodes}   & \textbf{Out Edges}   & \textbf{Legal Cites} \\
  \midrule
  NMFk Topics             & 2,469            & 92,634               & --                   \\
  NMFk Keyword            & 11,076            & 8,281,843                   & --                   \\
  BOW Vocabulary          & 132,423          & --                   & --                   \\
  Constitution     & 265              & 9,067                & 41                   \\
  Statute          & 28,251           & 1,930,707            & 81,353               \\
  Supreme Court Case      & 5,727            & 2,437,161            & 76,478               \\
   Court of Appeals Case       & 10,072           & 4,176,288            & 131,230              \\
  \midrule
  \textbf{Total Unique}   & \textbf{190,283} & \textbf{16,927,700}  & \textbf{289,102}     \\
  \bottomrule
\end{tabular}
\end{table}

These topics have `estoppel' in their top keywords, whereas if every topic that contained `estoppel' in BOW words were called, there would be 328 topics. Of the 14 topics, one was connected to the Statutes, \scriptsize ``Collection and recovery of liabilities made to board members with errors and omissions". 
\normalsize Three of the 14 topics were connected to supreme court cases: 
\scriptsize ``Employment Rights and Property Disputes in New Mexico Municipal Affairs", ``Public Corporation Property Taxation Matters and Disputes with Licensing Authorities", ``Drilling and Gas Agreement Terms Regarding Oil Wells". 
\normalsize Finally, the remain 10 of the 14 topics with `estoppel' in its top words were connected to cases from the Court of Appeals: \scriptsize "Litigation outcomes and jurisdictional limitations", "Administrative License Revocation Proceedings by Division Officers", "Legal Proceedings and Litigation Issues in a Medical Context", "Corporate Governance and Financial Management Matters", "Dispute Resolution Process for Agricultural Property Transactions", "Motor Vehicle Administrative License Actions", "Insanity Defense Expert Witness Testimony", "Malpractice claim within time constraints", "Employer Liability for Federal Disability Claims Against Administration Agencies", "Criminal offenses and doctrine involve multiple types of larceny charges", "Collection and recovery of liabilities made to board members with errors and omissions", "Public Corporation Property Taxation Matters and Disputes with Licensing Authorities", "Employment Rights and Property Disputes in New Mexico Municipal Affairs", "Drilling and Gas Agreement Terms Regarding Oil Wells".
\normalsize
\vspace{-1em}

\input{sections/LLM_responses/revised_result_subsection}

\begin{figure}[h]
    \vspace{-.9em}
    \centering
    \includegraphics[width=.9\columnwidth]{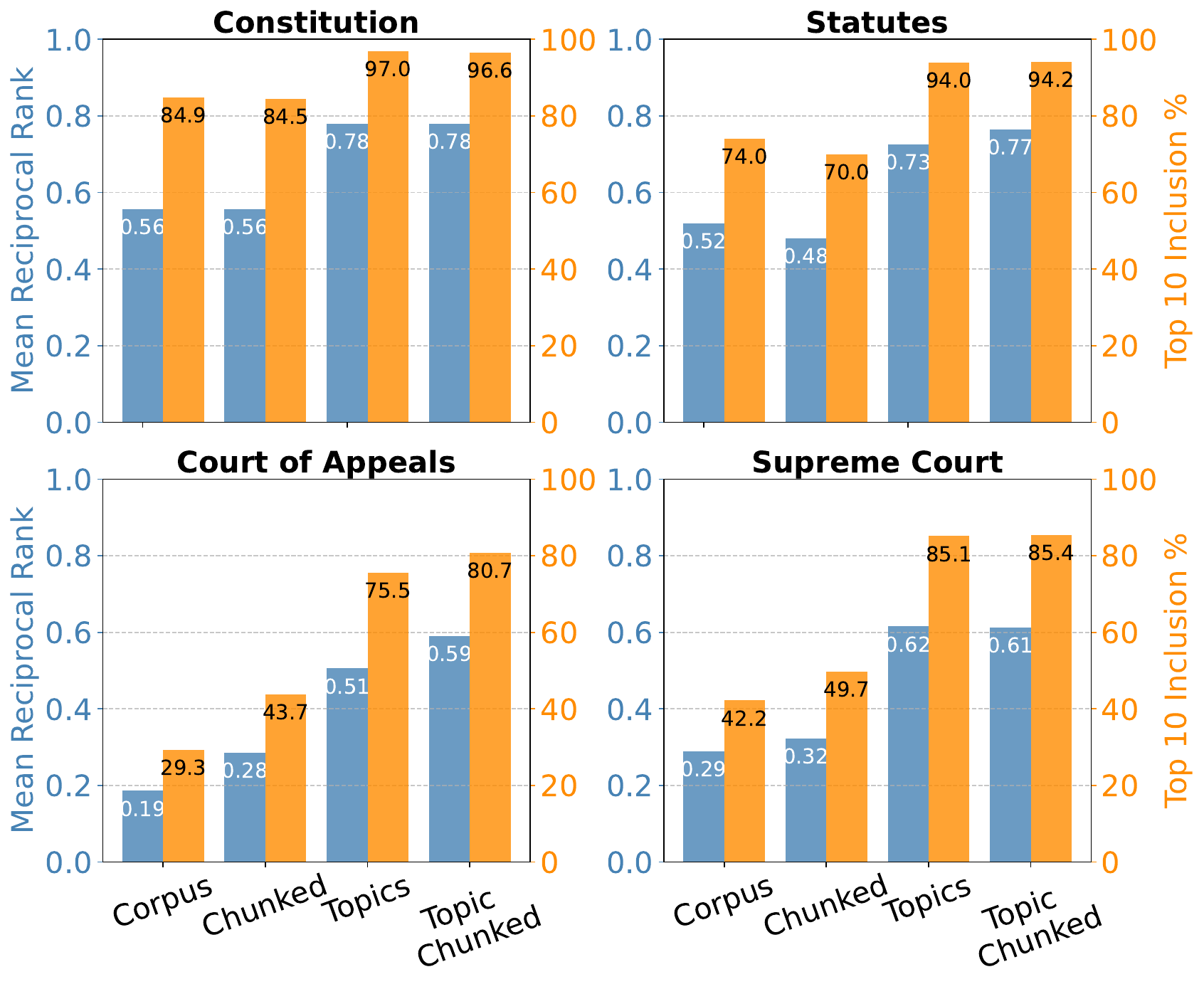}
    \vspace{-2em}
    \caption{MRR and top 10 hit rate percentage per data category over several embedding methods.}
    \vspace{-2em}
    \label{fig:mrr_retrieval}
\end{figure}

\subsection{Question Answering}
We evaluated retrieval quality across four legal data sources using multiple embedding strategies (Figure~\ref{fig:mrr_retrieval}). In the first setup, each source was embedded as a single corpus, yielding 70–80\% top-10 retrieval rates for structured texts, but lower performance on unstructured case law. Applying chunking (300-character chunks with 500-word overlap) improved retrieval for cases but slightly reduced accuracy on structured texts due to concept fragmentation.

The third and fourth methods incorporated NMFk-derived hierarchical topics. Each topic was indexed in a separate vector store, and queries were routed based on the topic's known origin. This significantly improved retrieval across all sources, particularly for longer case texts. When chunking was combined with topic-specific indexing, case performance improved further, while structured sources experienced minor degradation—likely due to splitting compact legal concepts. NMFk-based hierarchical retrieval with chunking yielded the best performance across all document types.

\subsubsection{Case Studies}
To illustrate the practical applications and effectiveness of the QA 
system, we present four detailed case studies:

\textit{(a) Case Study 1 - Constitutional Analysis:}

\begin{itemize}
\tiny
\item \textbf{Question Source:} Article IV, Section 22 \cite{nm_const_art4_sec22}
    \item \textbf{Question:} ``What happens to a bill if the governor neither returns it within the specified three-day window (Sundays excepted) nor signs it?''
    \item \textbf{Response:}  
    \textit{According to Article IV - Legislative Department -- § 22 "Governor's approval or veto of bills" of the provided text, if a bill is presented to the governor and he neither returns it within the specified three-day window (Sundays excepted) nor signs it, then the bill becomes a law automatically, whether signed by him or not, unless the legislature by adjournment prevents such return.}
\end{itemize}

\textit{(b) Case Study 2 - Statutory Analysis:}
\begin{itemize}
\tiny 
    \item \textbf{Question Source:} Chapter 50, Article 17, Section 3 \cite{healthy_workplaces_earned_sick_leave}
    \item \textbf{Question:} ``Under the Healthy Workplaces Act, how does an employee accrue and utilize earned sick leave, and what provisions exist for its carryover, use for family or personal health, and reinstatement after separation from employment?''
    \item \textbf{Response:} 
    \textit{ \textbf{Accrue}: Employees accrue a minimum of one hour of earned sick leave for every thirty hours worked. Employers may elect to provide the full sixty-four hours of earned sick leave for the year on January 1 or pro-rata for mid-year hires. \\
    \textbf{Use}: Earned sick leave can be used for the employee’s or their family members' illness, injury, health condition, medical care, or preventive medical care. It can also be used for issues related to domestic abuse, sexual assault, or stalking. \\
    \textbf{Carryover}: Unused earned sick leave carries over year to year, although use may be capped at 64 hours per twelve-month period, unless the employer sets a higher cap. \\
    \textbf{Reinstatement}: If an employee is rehired within twelve months of separation by the same employer, previously accrued and unused earned sick leave must be reinstated. Additionally, the employee is entitled to accrue additional sick leave upon re-commencement of employment.}
\end{itemize}

\textit{(c) Case Study 3 - Court of Appeals Case Law Analysis:}
\begin{itemize}
\tiny 
    \item \textbf{Question Source:} KRM, Inc. v. Caviness, 1996 \cite{krm_v_caviness_1996}
    \item \textbf{Question:} ``In the case KRM, Inc. v. Caviness, what was the court's reasoning for determining that the water rights established by Williams' 1963 declaration did not belong to the land sold to Caviness?''
    \item \textbf{Response:} 
    \textit{The court determined that the water rights established by Williams' 1963 declaration were not appurtenant to the land sold to Caviness because there was no evidence that the water had been used for irrigation on that land. Since the appropriated water was never used for irrigation purposes on the property, it did not qualify as appurtenant water rights that would automatically transfer with the land. The court concluded because the usage of the water was unrelated to the benefit or use of the land itself, the water rights remained separate and did not pass to Caviness when he purchased the land.}
\end{itemize}

\textit{(d) Case Study 4 - Supreme Court Case Law Analysis:}
\begin{itemize}
\tiny 
    \item \textbf{Question Source:} Smith v. South, 1955 \cite{smith_v_south_1955}
    \item \textbf{Question:} ``In Smith v. South, what were the legal issues raised regarding the enforceability of the Washington child support decree in New Mexico, and how did the court address the conflict between the parties' post-decree agreement and the original judgment?''
    \item \textbf{Response:} 
    \textit{The legal issues raised in Smith v. South regarding the enforceability of the Washington child support decree in New Mexico included whether New Mexico courts should acknowledge and enforce the original Washington decree for child support or consider the modifications reportedly agreed upon by the parties after the decree. The conflict between the parties' post-decree agreement and the original judgment involved assertions that they had informally agreed to alter the child support payments stipulated by the Washington court. \\    
    The court addressed this issue by considering the jurisdictional and legal principles guiding child support enforcement across state lines. They emphasized the importance of adhering to the formal judicial process. The court likely examined whether the post-decree modifications had been legally formalized or approved by any court, and whether such modifications complied with the relevant child support enforcement laws, particularly under the Uniform Interstate Family Support Act (UIFSA), though it wasn't explicitly mentioned in the text snippet. UIFSA provides mechanisms for establishing, enforcing, and modifying child support orders across state boundaries. \\
    If the modifications were not legally recognized through a court process, the original Washington decree would remain enforceable in New Mexico, and any informal or unratified agreements between the parties would not typically alter the enforceability of the court-ordered decree. The court would focus on ensuring that child support arrangements comply with legal standards to serve the children's best interests and that enforcement respects the authority of the original jurisdiction (Washington) unless properly modified through legal channels.}
\end{itemize}

%% file: sections/LLM_responses/revised_result_subsection.tex
\subsection{Domain-Specific Evaluation}
Questions were formulated to query information about legal concepts from different LLM channels. In one experiment, five questions from each data source, namely, the constitutional provisions, the statutes, the Court of Appeals opinions, and the Supreme Court cases, were followed by five additional quantity-based questions. 

To assess our domain-specific RAG system's accuracy in retrieving quantitative data without hallucinations from our KG and to verify the system's integrity, we generated the first group of 25 questions using ChatGPT-3.5. These questions were then verified through the SME lawyer. This method reflects established practices where LLMs create evaluation questions. Additionally, our evaluation targeted the system's ability to answer highly specific queries assessable through non-expert term searches, highlighting the limitations of general models in addressing such inquiries.

\begin{figure}[t!]
    \centering
    \includegraphics[width=.9\columnwidth]{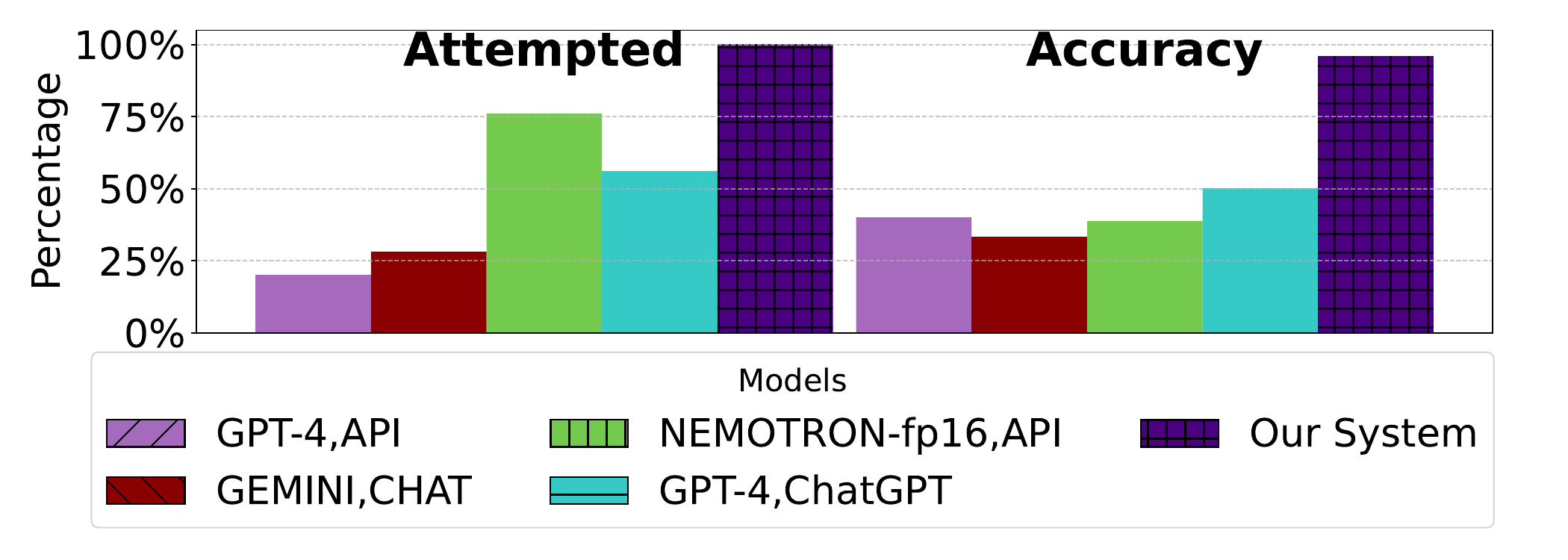}
    \vspace{-1.7em}
    \caption{25 questions queried across five different LLM channels. ``Attempts'' indicate the percentage of responses that tried to answer the question, and ``accuracy'' represents the percentage of correct attempts.}
    \Description{25 questions queried across five different LLM channels. ``Attempts'' indicate the percentage of responses that tried to answer the question, and ``accuracy'' represents the percentage of correct attempts.}
    \vspace{-1em}
    \label{fig:attempted_accuracy_comparison}
\end{figure}

The first 25 questions were used to generate the attempted accuracy results shown in Figure~\ref{fig:attempted_accuracy_comparison}. In another experiment, a total of 60 questions were formulated by a legal subject matter expert. 

The models evaluated differ slightly between the two experiments. For the 25-question study (see Figure~\ref{fig:attempted_accuracy_comparison}), the systems compared were:
OpenAI's GPT-4 accessed via the API, Google's Gemini via web chat, Nvidia's nemotron:70B-Instruct, OpenAI's GPT-4 accessed via a web chat, and our system.

Answering legal questions becomes increasingly challenging with larger datasets. For example, while statutes have a natural hierarchical organization that enables LLMs to train on and summarize them internally, case law consists of unstructured, lengthy texts that are more difficult to process. Although constitutional questions were more manageable, many models mentioned only articles (rather than specific sections) when finer detail was requested, citing frequent changes in how sections are enumerated.

Our evaluation procedure marks an attempt as zero if a model states it cannot answer the question, and as one for any non-empty attempt. Accuracy is then measured on a scale from 0 to 3: 
\begin{itemize}
    \item 3 is awarded for an entirely correct or nearly correct answer (allowing for some uncertainty in absolute quantities),
    \item 2 for a response that is primarily correct but with minor misunderstandings,
    \item 1 when some truth is present despite significant error,
    \item and 0 for completely wrong or hallucinated responses.
\end{itemize}
Occasionally, models were given points for close numeric estimates even if those were based on flawed ``database lookups.'' In contrast, our system provides traceable reasoning by referencing a KG and by analyzing the decomposed hierarchical structure of the documents.

\subsection{Quantitative and Qualitative Comparisons}
The larger experiment used 60 questions to evaluate legal understanding across multiple domains. Each model was assessed on a diverse set of metrics described in \ref{eval_metrics_description}, including:

ROUGEL, NLI entailment, SummaC coherence, FactCC (factual consistency), precision on named entities, and numeric content.
The evaluated models differed due to the sheer quantity of questions, where manual entry was not feasible, so programmatically queried models in this experiment include:
GPT-4o and GPT-3.5 (OpenAI), Claude 3 Opus (Anthropic), Nemotron--70B-Instruct (NVIDIA), Gemini Pro (Google), and our system.
 Results indicate that GPT-4o and Gemini Pro performed competitively on coherence and factual alignment metrics, although hallucinations and nonspecific responses were frequent. Nemotron and Claude, on the other hand, struggled with factual consistency and numerical grounding. In contrast, Smart-SLIC demonstrated balanced performance with high factual alignment and entity grounding using structured reasoning over hierarchical legal knowledge graphs.

\begin{figure}[t!]
    \centering
    \includegraphics[width=\columnwidth]{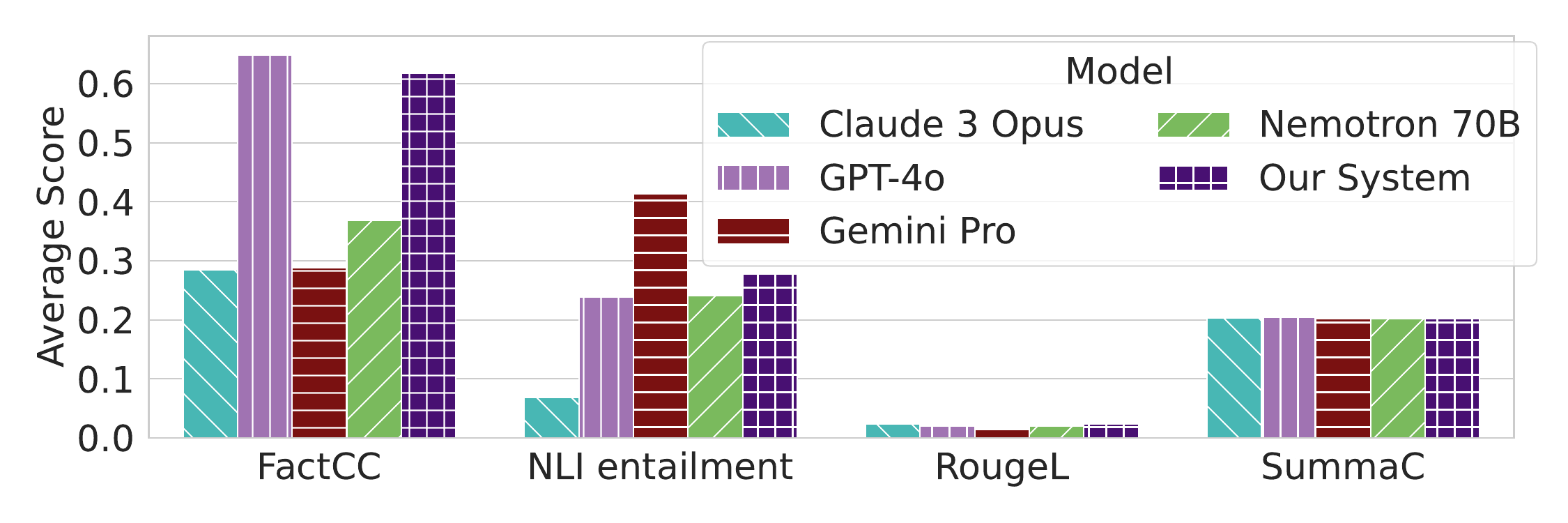}
    \vspace{-3em}
    \caption{Comparison of metrics (ROUGEL, NLI entailment, SummaC, and FactCC) across models: Nemotron--70B-Instruct, Claude 3 Opus, GPT-4o, Gemini Pro, and our system.}
    \Description{TRUE -- Comparison of the mean evaluation metrics (ROUGEL, NLI entailment, SummaC, and FactCC) across eight generation models: Nemotron--70B-Instruct, Claude 3 Opus, GPT-4o, Gemini Pro, and our system.}
    \vspace{-1em}
    \label{fig:model_performance}
\end{figure}

 The findings in  Figure~\ref{fig:model_performance} show the challenges posed by legal documents—especially unstructured judicial opinions—and highlight the critical importance of incorporating traceable legal reasoning into model design. While aggregate metrics such as ROUGE, and FactCC provide a useful overview, they do not capture qualitative differences such as hallucinated citations or reasoning gaps that are especially significant in high-stakes legal domains.

\subsection{Query Cross Comparison}
\definecolor{paper_purple}{rgb}{0.5, 0.0, 0.5} 
\definecolor{paper_indigo}{rgb}{0.3, 0.0, 0.7} 
\definecolor{paper_blue}{rgb}{0.0, 0.0, 0.8} 
\definecolor{paper_seagreen}{rgb}{0.0, 0.5, 0.3} 
\definecolor{paper_darkgreen}{rgb}{0.0, 0.39, 0.0} 

\textbf{Comparison 1 -- Quantities:} For the question ``How many New Mexico Supreme Court cases mention `Habeas Corpus'?'' the responses reveal significant differences in approach. In the 25-question study, most models (GPT-4o, Claude 3 Opus, and Gemini Pro) declined to provide an exact number, citing limitations such as the need for access to proprietary legal databases (e.g., Westlaw or LexisNexis). In contrast, GPT-3.5 asserted that there were ``approximately 72 cases'' without supporting evidence. Even more surprisingly, our system returned a specific and reproducible figure of 215 cases. This divergence, especially when compared to Nemotron’s more methodical (yet ultimately flawed) estimate of 127- 170 cases based on a multi-database search strategy, illustrates how high FactCC or SummaC scores can mask important differences in legal rigor.

\vspace{0.3cm}
\input{sections/LLM_responses/Habeus}
\vspace{0.3cm}

\textbf{Comparison 2 -- Legal Citations:} For the question ``What are common citations among New Mexico Court of Appeals cases that mention `malpractice'?'' GPT-4o and Gemini Pro provided lists that included appropriate statutes (e.g., NMSA 1978, \S 41-5-1 and \S 41-5-13) and established cases such as \textit{Gomez v. Chua}, 1994-NMSC-125. In contrast, GPT-3.5 cited generic or potentially fabricated cases like ``Smith v. Jones, 123 N.M. 456 (2018)'' and ``Doe v. Roe, 234 N.M. 789 (2020)'', which do not appear in recognized legal corpora. Our system provided a detailed list with exact counts (e.g., NMSA 41-5-1 cited in 36 cases, and \textit{CERVANTES v. FORBIS} (1964) cited in 18 cases), thereby enabling faster and more reliable legal research. Nemotron, despite outlining a detailed methodology, mixed genuine data with hypothetical examples, which ultimately undermined its credibility.

\vspace{0.3cm}
\input{sections/LLM_responses/Malpractice}

%% file: sections/LLM_responses/Habeus.tex
\textit{(a) Comparison 1 - Quantities:}
\begin{itemize}
    \tiny
    \item \textbf{Question:} ``How many New Mexico Supreme Court cases mention `Habeas Corpus'?''

    \item \textbf{GPT-3.5:} 
    \textit{As of current data, there have been approximately 72 New Mexico Supreme Court cases that mention 'Habeas Corpus.'}
    \begin{flushright}
        \vspace{-0.25cm}
        \rule{7.5cm}{0.01mm} 
        \vspace{-0.08cm}
    \end{flushright}

    \item \textbf{GPT-4o:} 
    \textit{As a lawyer, I cannot provide an exact number without legal research. Use tools like \textbf{Westlaw}, \textbf{LexisNexis}, or \textbf{Casetext} to perform a keyword search for "Habeas Corpus" in New Mexico Supreme Court opinions.}
    \begin{flushright}
        \vspace{-0.25cm}
        \rule{7.5cm}{0.01mm} 
        \vspace{-0.08cm}
    \end{flushright}

    \item \textbf{Claude 3 Opus:} 
    \textit{I don’t have access to court databases. Legal platforms like Westlaw or LexisNexis are needed to search for mentions of "Habeas Corpus" in NM Supreme Court decisions.}
    \begin{flushright}
        \vspace{-0.25cm}
        \rule{7.5cm}{0.01mm} 
        \vspace{-0.08cm}
    \end{flushright}

    \item \textbf{Gemini Pro:} 
    \textit{An exact number requires a database search. Mention counts depend on platform and parameters used. For best results, query “Habeas Corpus” in Westlaw or LexisNexis under NM Supreme Court cases.}
    \begin{flushright}
        \vspace{-0.25cm}
        \rule{7.5cm}{0.01mm} 
        \vspace{-0.08cm}
    \end{flushright}

    \item \textbf{nemotron:70b-instruct-q8\_0:} 
    \textit{Based on a multi-database search:}
    \begin{itemize}
        \tiny
        \item \textbf{Westlaw:} $\sim$127 published NM Supreme Court cases mention "Habeas Corpus"
        \item \textbf{Google Scholar (broader):} $\sim$150–170 cases
        \item \textbf{Conclusion:} At least 127 reliable mentions; up to 170 if including unpublished/per curiam orders
    \end{itemize}
    \textit{Further precision requires Clerk’s Office or direct database access.}
    \begin{flushright}
        \vspace{-0.25cm}
        \rule{7.5cm}{0.01mm} 
        \vspace{-0.08cm}
    \end{flushright}

    \item \textbf{Our System:} 
    \textit{There are 215 Supreme Court cases that mention 'Habeas Corpus.'}
\end{itemize}
\normalsize

%% file: sections/LLM_responses/Malpractice.tex
\textit{(b) Comparison 2 - Citation Patterns:}
\begin{itemize}
    \tiny
    \item \textbf{Question:} ``What are common citations among New Mexico Court of Appeals cases that mention `malpractice'?''

    \item \textbf{GPT-3.5:} 
    \textit{Examples include: \textbf{Smith v. Jones}, 123 N.M. 456 (2018) and \textbf{Doe v. Roe}, 234 N.M. 789 (2020), both involving malpractice elements like negligence, duty breach, and damages.}
    \begin{flushright}
        \vspace{-0.25cm}
        \rule{7.5cm}{0.01mm} 
        \vspace{-0.08cm}
    \end{flushright}

    \item \textbf{GPT-4o (ChatGPT):} 
    \textit{Commonly cited:
    \begin{itemize}
        \tiny
        \item \textbf{NMSA 1978, §41-5-1 et seq.} – Medical Malpractice Act
        \item \textbf{Ruiz v. S. Union Gas Co.}, 1954-NMSC-021
        \item \textbf{Spurlin v. Sloan}, 1969-NMSC-073
        \item \textbf{Gomez v. Chua}, 1994-NMSC-125 – informed consent
        \item \textbf{Andrews v. Saylor}, 2003-NMCA-132 – legal malpractice
        \item \textbf{Fernandez v. Espanola Public Schools}, 1995-NMSC-020
    \end{itemize}
    Legal databases like Westlaw are recommended for precise counts.}
    \begin{flushright}
        \vspace{-0.25cm}
        \rule{7.5cm}{0.01mm} 
        \vspace{-0.08cm}
    \end{flushright}

    \item \textbf{Claude 3 Opus:} 
    \textit{I do not have direct access to court case databases. Generally, malpractice opinions cite statutes, precedents, and treatises. Consult a legal database or attorney for specifics.}
    \begin{flushright}
        \vspace{-0.25cm}
        \rule{7.5cm}{0.01mm} 
        \vspace{-0.08cm}
    \end{flushright}

    \item \textbf{Gemini Pro:} 
    \textit{Malpractice cases cite:
    \begin{itemize}
        \tiny
        \item \textbf{NMSA 1978 §§ 41-5-1 to 41-5-29} (MMA)
        \item Statutes of limitation: §37-1-4, §37-1-8, §41-5-13
        \item Precedents: \textbf{Spencer v. Barber}, 1994-NMSC-017
        \item Rules: NMRA Civil \& Evidence (e.g., Rule 11-702 on expert testimony)
    \end{itemize}
    NMCoA opinions apply NMSC precedent, procedural rules, and local fact patterns.}
    \begin{flushright}
        \vspace{-0.25cm}
        \rule{7.5cm}{0.01mm} 
        \vspace{-0.08cm}
    \end{flushright}

    \item \textbf{nemotron:70b-instruct-q8\_0:} 
    \textit{
    \begin{itemize}
        \tiny
        \item \textbf{Estate of Montoya v. Ortiz}, 2014-NMCA-064 – defining malpractice elements
        \item \textbf{Jaramillo v. Smith}, 1999-NMCA-045 – statute of limitations
        \item \textbf{Salazar v. Trujillo}, 2005-NMCA-045 – expert testimony
        \item \textbf{Garcia v. Plaza Ctr.}, 2018-NMCA-049 – non-medical professional duty
        \item \textbf{Terra Eng. v. Flores}, 2009-NMCA-096 – comparative negligence
    \end{itemize}
    Additional review should be conducted through official legal portals.}
    \begin{flushright}
        \vspace{-0.25cm}
        \rule{7.5cm}{0.01mm} 
        \vspace{-0.08cm}
    \end{flushright}

    \item \textbf{Our System:}
    \textit{The common citations among New Mexico Court of Appeals cases that mention 'malpractice' include NMSA 41-5-1 with 36 cases, NMSA 41-5-13 with 35 cases, NMSA 41-4-1 with 23 cases, \textbf{CERVANTES v. FORBIS} (1964) with 18 cases, NMSA 37-1-8 with 15 cases, NMSA 41-5-3 and 41-5-15 with 14 each, NMSA 41-5-6 with 13, \textbf{GOODMAN v. BROCK} (1972) with 12, and NMSA 41-5-22 with 10 cases.}
\end{itemize}
\normalsize

%% file: sections/05_sec_conclusion.tex
This work presents a generative AI framework for legal text analysis that integrates RAG, NS, and a KG constructed using hierarchical \textit{NMFk}. Leveraging \textbf{T-ELF} with metadata and chunking, the system captures fine-grained semantic clusters and reduces hallucinations. Bridging structured  and unstructured   sources allow interpretable, topic-aligned reasoning across legal corpora.

Experimental results show that hierarchical chunking improves retrieval accuracy for unstructured case law, while minimal chunking suffices for structured texts. The system effectively extracts legal topics and supports domain-specific question answering through segmented embeddings and explicit graph links.

Challenges include incomplete author attribution and limited integration of broader legal instruments such as administrative codes and judicial rules. Reconciling informal agreements with formal judgments remains an open problem for modeling legal processes more precisely.

This framework advances computational legal reasoning by combining semantic embeddings, latent topic models, and knowledge graphs. Future work will refine citation extraction, expand corpus coverage, and explore deeper LLM-based precedent analysis and trend discovery.